\PassOptionsToPackage{table,xcdraw,dvipsnames}{xcolor}

\documentclass[10pt,twocolumn,letterpaper]{article}
\usepackage[accsupp]{axessibility}  

\usepackage{algorithm}      
\usepackage{algpseudocode} 

\usepackage{microtype}

\usepackage{cvpr}            

\usepackage{graphicx}
\usepackage{amsmath}
\usepackage{amssymb}
\usepackage{booktabs}

\definecolor{cvprblue}{rgb}{0.21,0.49,0.74}
\usepackage[pagebackref,breaklinks,colorlinks,allcolors=cvprblue]{hyperref}

\usepackage{url}            
\usepackage{amsfonts}       
\usepackage{nicefrac}       
\usepackage{microtype}      

\usepackage{makecell}
\usepackage{amsthm}
\usepackage{bm}
\usepackage[table,xcdraw]{xcolor}
\usepackage{multicol}
\usepackage{colortbl}
\usepackage{multirow}
\usepackage{enumitem}
\usepackage{bbding}
\usepackage{color}

\usepackage{algorithm}
\usepackage{listings}
\usepackage{natbib}

\usepackage{etoolbox}
\makeatletter
\AfterEndEnvironment{algorithm}{\let\@algcomment\relax}
\AtEndEnvironment{algorithm}{\kern2pt\hrule\relax\vskip3pt\@algcomment}
\let\@algcomment\relax
\newcommand\algcomment[1]{\def\@algcomment{\footnotesize#1}}
\renewcommand\fs@ruled{\def\@fs@cfont{\bfseries}\let\@fs@capt\floatc@ruled
  \def\@fs@pre{\hrule height.8pt depth0pt \kern2pt}%
  \def\@fs@post{}%
  \def\@fs@mid{\kern2pt\hrule\kern2pt}%
  \let\@fs@iftopcapt\iftrue}
\makeatother

\definecolor{ForestGreen}{RGB}{34,139,34}

\renewcommand{\paragraph}[1]{\medskip\noindent\textbf{#1.~}}

\newcommand{\modelname}{LaPR}

\newcounter{qcounter}

\def\confName{CVPR}
\def\confYear{2026}
\def\modelname{\textbf{\texttt{LaPR}}} 

\title{\emph{Love Me, Love My Label}:\\Rethinking the Role of Labels in Prompt Retrieval for Visual In-Context Learning}

\author{
Tianci Luo$^{1}\thanks{These authors contributed equally to this work.}$ , \ 
Haohao Pan$^{3*}$, \ 
Jinpeng Wang$^{2}\thanks{Corresponding author.}\;$, \ 
Niu Lian$^{2}$, \
Xinrui Chen$^{1}$, \\
Bin Chen$^{2}$ , \ 
Shu-Tao Xia$^{1}$, \ 
Chun Yuan$^{1}$ \\
{\normalsize $^1$Tsinghua Shenzhen International Graduate School, Tsinghua University} \\
{\normalsize $^2$Harbin Institute of Technology, Shenzhen} \\
{\normalsize $^3$School of Computer Science and Engineering, Northeastern University} \\
{\normalsize \tt ltc25@mails.tsinghua.edu.cn \qquad \small \tt panhh@mails.neu.edu.cn \qquad 
 \small\ \tt wangjp26@gmail.com} \\
}

\begin{document}
\maketitle
\begin{abstract}

Visual in-context learning (VICL) enables visual foundation models to handle multiple tasks by steering them with demonstrative prompts. 
The choice of such prompts largely influences VICL performance, standing out as a key challenge.
Prior work has made substantial progress on prompt retrieval and reranking strategies, but mainly focuses on prompt images while overlooking labels. 
We reveal these approaches sometimes get visually similar but label-inconsistent prompts, which potentially degrade VICL performance. 
On the other hand, higher label consistency between query and prompts preferably indicates stronger VICL results. 
Motivated by these findings, we develop a framework named \modelname{} (\textbf{L}abel-\textbf{a}ware \textbf{P}rompt \textbf{R}etrieval), which highlights the role of labels in prompt selection. 
Our framework first designs an image–label joint representation for prompts to incorporate label cues explicitly. 
Besides, to handle unavailable query labels at test time, we introduce a mixture-of-expert mechanism to the dual encoders with query-adaptive routing. 
Each expert is expected to capture a specific label mode, 
while the router infers query-adaptive mixture weights and helps to learn label-aware representation.
We carefully design alternative optimization for experts and router, with a VICL performance-guided contrastive loss and a label-guided contrastive loss, respectively. 
Extensive experiments show promising and consistent improvement of \modelname{} on in-context segmentation, detection, and colorization tasks. 
Moreover, \modelname{} generalizes well across feature extractors and cross-fold scenarios, suggesting the importance of label utilization in prompt retrieval for VICL. 
Code is available at https://github.com/luotc-why/CVPR26-LaPR.

\end{abstract}    
\section{Introduction}
\label{sec:intro}

\begin{figure}
    \centering
    \includegraphics[width=1\linewidth]{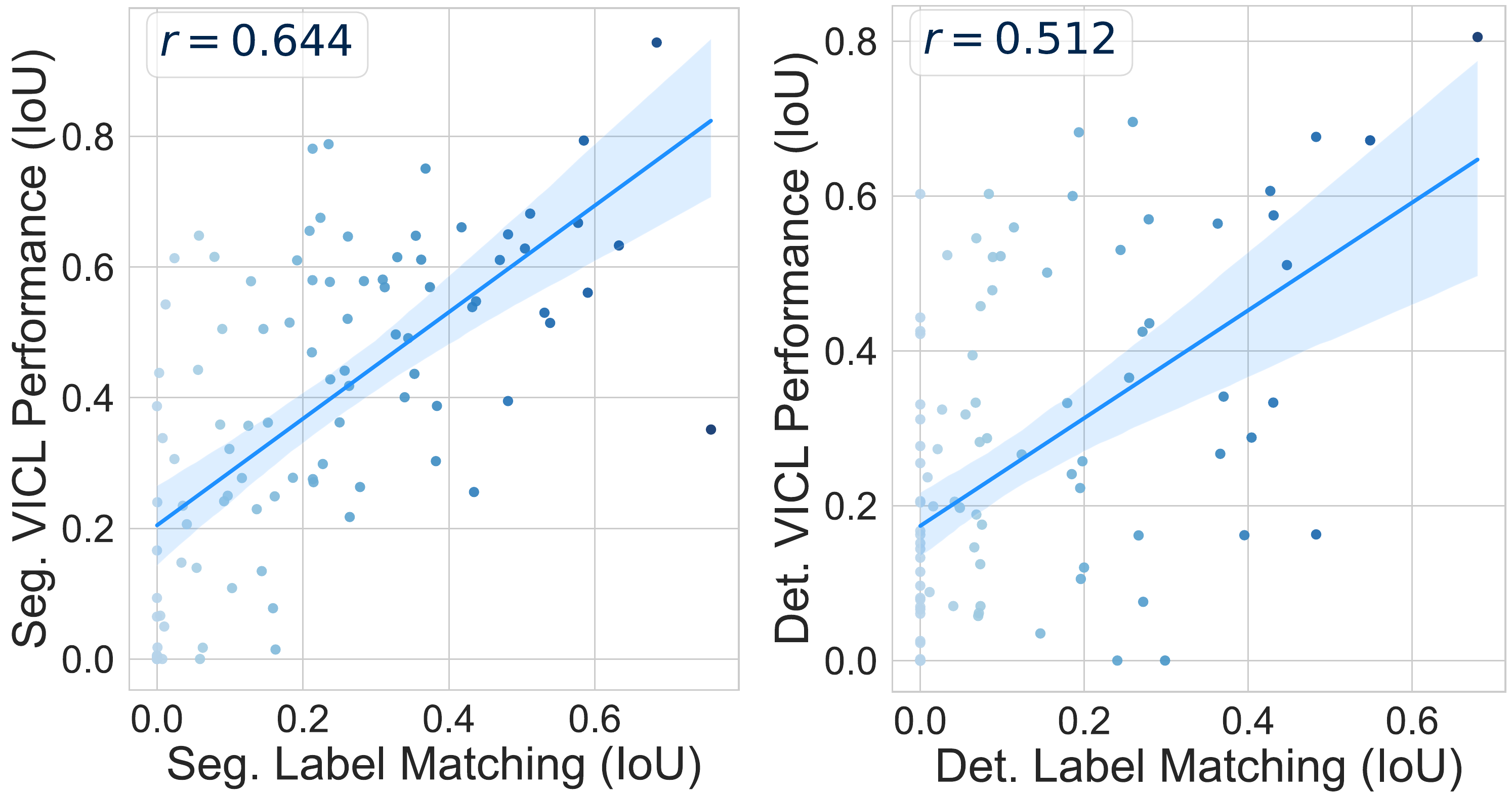}
    \caption{We randomly sample \(100\) query-prompt training pairs constructed by SupPR \cite{SupPR}. 
    We compute the label matching consistency and VICL performance to investigate their correlation.}
    \label{fig:intro_1}
\end{figure}

\begin{figure*}[t]  
\centering       
\includegraphics[width=1\textwidth]{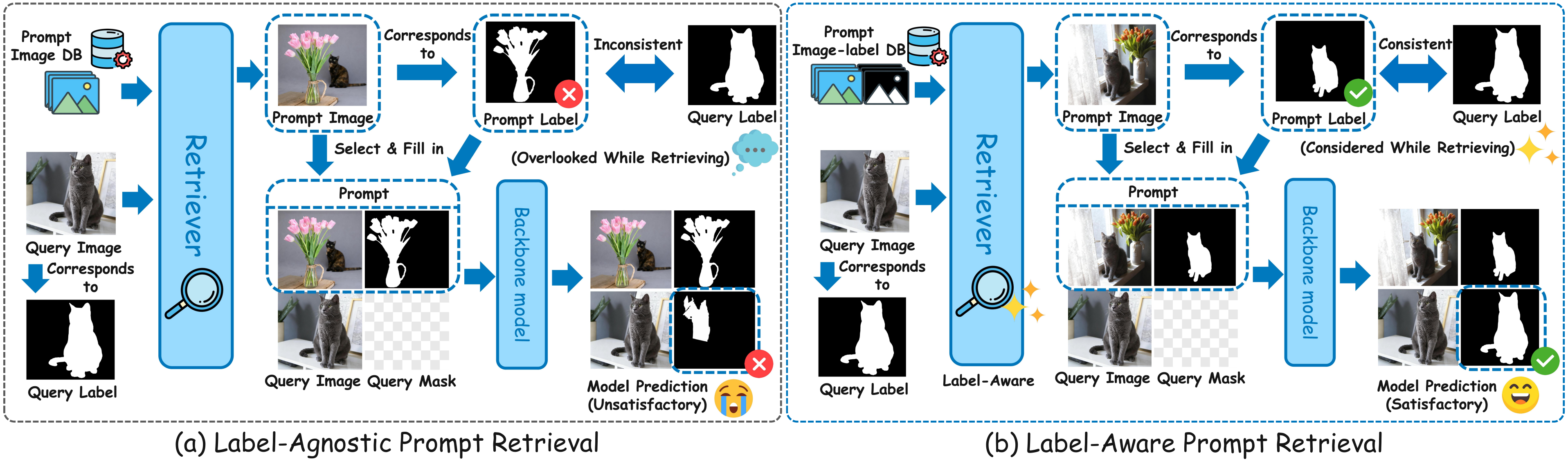} 
\caption{Prompt retrieval paradigms. 
(a) Label-agnostic pipelines rely on image similarity and may yield label-inconsistent prompts and disturb inference. 
(b) \modelname{} considers both image similarity and \emph{label consistency} for prompt retrieval, retrieving more relevant prompts.}
\label{fig:intro_2}  
\end{figure*}

Large foundation models~\cite{brown2020language,touvron2023llama,alayrac2022flamingo} are favoured in many user-centric applications for their strong in-context learning (ICL) capacity, where different demonstrative prompts enable a model to handle various tasks. 
Following the emergence of a series of vision backbones~\cite{fang2024explore,MAE-VQGAN,wang2023images,segGPT,LVM}, Visual ICL (VICL) has also gained increasing attention from the computer vision community. 
A milestone is MAE-VQGAN \cite{MAE-VQGAN}, which typically formulates VICL as pixel inpainting. The input image is organized in a $2 \times 2$ grid: the top half contains the prompt image and its pixel-format label, the bottom-left cell holds the query image, and the lower-right cell is masked out. The model's task is to reconstruct this masked region and yield the label prediction to the query.

Existing studies~\cite{25SunExplore,SupPR,liu2021makes,paibu_nlp} dive into effective factors to improve ICL, where prompt selection plays a critical role.
More recent works have focused on retrieving better prompts to enhance VICL performance.
For instance, \citet{SupPR} defined positive and negative prompt pairs based on their contribution to VICL inference performance and introduced contrastive learning into retriever optimization, while \citet{Partial2Global,wu2025towards} reformulated prompt selection as a reranking process and achieved notable improvements through score-based prediction.
However, at inference, only the query image is observable, and for input symmetry, the prompt side typically omits labels, which forfeits useful information. 
As shown in \Cref{fig:intro_2}(a), for a query whose subject is a cat, the system might retrieve a prompt image containing both a cat and a flower but annotated with the label ``flower", leading to erroneous in-context predictions. 
Interestingly, as illustrated in \Cref{fig:intro_1}, we observe that among prompts with relevant images, the label consistency with the query is positively correlated to the VICL performance. It motivates us to make better use of label information in prompt retrieval for VICL. 
As shown in \Cref{fig:intro_2}(b), our goal is to enhance query-prompt label consistency in prompt retrieval, encouraging more accurate and reliable in-context predictions.

To this end, we propose a new paradigm for prompt retrieval called \modelname{} (\textbf{L}abel-\textbf{a}ware \textbf{P}rompt \textbf{R}etrieval). 
We design label-aware strategies for the prompt and query, respectively. 
For the prompt, we explicitly inject label information into the representation by fusing image and label features into joint representation. 
For the query, more challengingly, to tackle the unavailable label at test time, we adopt mixture-of-expert designs for the dual encoders, aiming to perceive and adapt to the implicit (\ie, unknown) query label. 
Specifically, each expert in the prompt and the query encoder is designated to capture a distinct mode (\eg long tail, horn, sharp beak), and the query-dependent router infers soft mixture weights to entail the implicit query label, obtaining adaptive query and prompt embeddings, as well as similarity scores adaptive to the implicit query label. 
In particular, to decouple the roles of the experts and the router, we adopt alternating optimization with two successive updates per mini-batch. The expert step learns via a performance-guided contrastive objective to strengthen mode-specific encodings. The router step optimizes a label-guided objective to align query-adaptive modes and adds the load-balancing regularizer to avoid expert under-utilization.

Extensive experiments substantiate the effectiveness of \modelname{} framework.
\modelname{} consistently achieves state-of-the-art performance across foreground segmentation, single-object detection, and colorization tasks.
Furthermore, it exhibits robustness cross feature extractors generalization and remarkable transferability under cross-folder settings.
These results underscore the pivotal role of label information in VICL prompt selection, and establish a new foundation for label-aware prompt retrieval in VICL.

To Sum up, we make main contributions as follows:

\setlist{nolistsep}
\begin{itemize}[leftmargin=1em]

\item We present the first \emph{label-aware prompt retrieval} in VICL, explicitly addressing label inconsistency and offering new conceptual insights into prompt selection. 

\item We inject prompt labels to create label-aware embeddings, and depict different modes with multiple experts on both sides. The query-specific router assigns mixture weights to experts, helping to yield an estimated query label and guide the extraction of query-relevant prompt information.

\item We decouple the roles of the experts and the router. The experts are trained to strength mode-specific representations, while the router is trained to infer mixture proportions over modes that align with the ground truth query label.

\item \modelname{} attains state-of-the-art results on foreground segmentation, single-object detection, and colorization tasks, and shows robust cross-backbone generalization and transferability under cross-folder scenario.

\end{itemize}

\section{Related Works}
\label{sec:related_works}

\subsection{In-Context Learning}
\label{subsec:related_work_a}
In-context learning (ICL) empowers large language models to learn task patterns and acquire task completion abilities by providing a few examples within the input prompt. This approach has seen extensive development \cite{brown2020language,alayrac2022flamingo} and application \cite{min2021metaicl,zhou2024visual} in natural language processing (NLP) and multi-modal fields.
Furthermore, the theoretical foundations of ICL have been rigorously validated \cite{von2023transformers}. Building upon its broad applicability and theoretical grounding, ICL has undergone significant advancements, particularly in improving example retrieval methods \cite{liu2021makes, rubin2021learning, wu2022self, guo-etal-2024-makes, peng2024revisiting, yan2026learning}. 

The emergence of a series of vision backbones \cite{segGPT,LVM,point_mae,fang2024explore} has further advanced the application of ICL in the visual domain, termed Visual In-Context Learning (VICL).
MAE-VQGAN \cite{MAE-VQGAN} and Painter \cite{wang2023painter} have demonstrated VICL across various tasks such as colorization and segmentation. Oorloff et al. \cite{applestable} propose an in-place attention mechanism for ICL paradigm, achieving notable results.
Building on these research, several improvements have been proposed. For example, InMeMo \cite{zhang2024instruct} improved VICL performance by introducing border noise, while Condenser \cite{condenser} and PromptHub \cite{luo2026prompthub} explored multi-prompts fusion to facilitate input tokens limitation.
Prompt-SelF~\cite{PromptSelF} demonstrates that the arrangement of prompts affects VICL behavior and incorporates a voting-based strategy to improve robustness.
In the realm of prompt retrieval, 
SupPR \cite{SupPR} leverages contrastive learning to enhance the retriever, whereas Partial2Global \cite{Partial2Global} reformulates the retrieval process from a reranking perspective. RH-Partial2Global~\cite{RH-Partial2Global} constructs stable candidate sets based on a jackknife conformal prediction strategy, while employing a covering design–based sampling scheme to achieve comprehensive and uniform retrieval.

Moreover, \citet{wang-etal-2023-label} investigated the role of label in information aggregation and prediction referencing within ICL. Although existing VICL retrievers focus on image part similarity, the prompt label information has been largely ignored. In this work, we make the first attempt to explicitly exploit prompt labels for a more effective retrieval.

\subsection{Mixture of Experts}
\label{subsec:related_work_c}
The underlying principle of Mixture of Experts (MoE) \cite{ShazeerMMDLHD17, masoudnia2014mixture} is to employ a collection of expert networks, each specializing in handling specific tasks or subsets of the input space. MoE has been widely applied across various fields \cite{scalingvision, FedMoE, Chen_2023_ICCV, Yu_2024_CVPR}. For instance, in the area of information retrieval, SA-MoE \cite{SA-MoE} leverages the MoE mechanism to enhance semantic feature representation in unsupervised cross-domain image retrieval, while DESIRE-ME \cite{DESIRE-ME} employs MoE to improve retrieval performance in cross-domain open-domain question answering. Moreover, ICL works as MOICL \cite{hong2024moicl} and MoD \cite{wang2024mixture} enhance performance by using MoE for expert-based prompt retrieval.
In this work, we encode mode-specific embeddings via specialized experts, with a query-adaptive router determines the mixture weights across modes to help realize label-aware embeddings.

\section{Methods} \label{sec:methods}

\begin{figure*}[t]
    \centering
    \includegraphics[width=\textwidth]{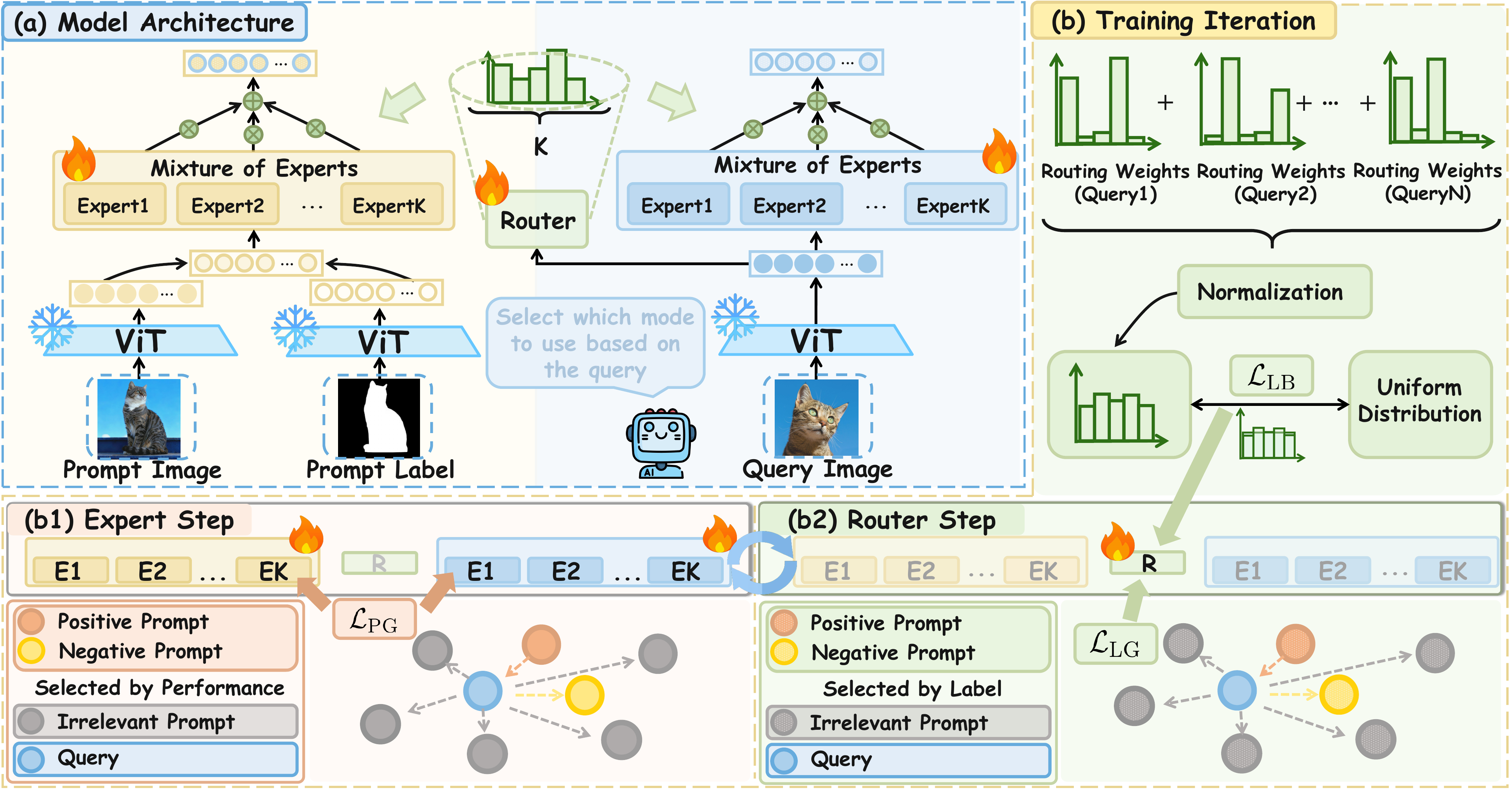}
    \vspace{-1.5em}
    \caption{Overview of \modelname{}.
    (\textbf{a}) \modelname{} architecture. Prompt labels are injected to form joint embeddings. 
    On both sides, experts produce mode specific features and a query conditioned router picks the matching mode and extracts its information, resulting in label-aware query embeddings and query-relevant prompt embeddings.
    (\textbf{b}) Training framework. Each mini-batch alternates an expert step (performance-guided contrastive learning, router fixed) and a router step (label-guided contrastive learning with load balancing, experts frozen).}
    \vspace{-0.5em}
    \label{fig:arc}
\end{figure*}

\subsection{Problem Formulation and Overview of \modelname{}} \label{subsec:overview}
Let $\mathcal{B}=\{(I_i^{p},L_i^{p})\}_{i=1}^{N}$ be a database of candidate prompt, where $I_i^{p}$ is a prompt image and $L_i^{p}$ is its label (e.g., depth map, segmentation mask, or colorized images).
At inference, a query image $x_q$ arrives with an \emph{unobserved} label $y_q$.  

VICL demonstrative prompt selection seeks the single best prompt \(c_q^{\star}\in\mathcal{B}\) for the query \(x_q\). A formal objective is
\begin{equation}
c_q^{\star}\in\arg\max_{(I_i^{p},L_i^{p})\in\mathcal{B}}
\mathcal{S}_{\text{task}}\big(\Psi\big(x_q,(I_i^{p},L_i^{p})\big),\,y_q\big).
\end{equation}
Here \(\Psi(x_q;(I_i^{p},L_i^{p}))\) denotes the prediction, \(\mathcal{S}_{\text{task}}\) is the task-specific metric, \(c_q^{\star}\) is the selected best prompt from \(\mathcal{B}\).

In practice we instantiate the selector by either prompt retrieval or prompt rerank within one formulation. For prompt retrieval we encode prompts \(z_i^I=f(I_i^{p})\in\mathbb{R}^{d}\) and the query \(u_q=f(x_q)\in\mathbb{R}^{d}\), compute similarity score \(s_{\mathrm{ret}}(q,i)=\sigma(u_q,z_i^I)\), and select \(i^{\star}=\arg\max_{i\in[N]} s_{\mathrm{ret}}(q,i)\) with \(c_q^{\star}=(I_{i^{\star}}^{p},L_{i^{\star}}^{p})\). For prompt rerank we are given a candidate pool \(\tilde{\mathcal{C}}_q=\{(I_{i_m}^{p},L_{i_m}^{p})\}_{m=1}^{M}\subset\mathcal{B}\) and evaluate a query-conditioned listwise scorer to obtain score vector \(\mathbf{s}_{\mathrm{re}}(q,\tilde{\mathcal{C}}_q)=g\!\big(x_q,\{I_{i_m}^{p}\}_{m=1}^{M}\big)\in\mathbb{R}^{M}\), then choose \(m^{\star}=\arg\max_{m\in[M]}[\mathbf{s}_{\mathrm{re}}(q,\tilde{\mathcal{C}}_q)]_{m}\) and set \(c_q^{\star}=(I_{i_{m^{\star}}}^{p},L_{i_{m^{\star}}}^{p})\), where \(f:\mathcal{X}\to\mathbb{R}^{d}\) is an image encoder, \(\sigma:\mathbb{R}^{d}\times\mathbb{R}^{d}\to\mathbb{R}\) is a similarity function, and \(g\) maps a query and an image list to a score vector in \(\mathbb{R}^{M}\).

However, the prompt label part $L^p_i$ that naturally accompanies each prompt is often discarded during selection, which may lead to label inconsistency and degraded VICL performance. 
To exploit label cues, we propose \modelname{}, which builds joint embeddings for prompts and introduces a mixture-of-expert mechanism to capture distinct mode-specific embeddings for both sides with a query-adaptive router.
The router infers mixture weights, which are then used to aggregate the query-relevant prompt embeddings and form label-aware query embeddings by combining modes to estimate the query label. 
Then we compute the label-aware retrieval score. 
(see \Cref{fig:arc}).

\subsection{Label-Aware Representation} \label{subsec:embeddings} 

We first encode the images with the feature extractor $f$, obtaining the embeddings. Concretely, for each prompt pair \((I_i^{p},L_i^{p})\) we compute \(z_i^{I}=f(I_i^{p})\in\mathbb{R}^{d}\) and \(z_i^{L}=f(L_i^{p})\in\mathbb{R}^{d}\), and for the query we set \(u_q=f(x_q)\in\mathbb{R}^{d}\).

\paragraph{Label-Aware Query Embeddings}

Drawing on the Mixture-of-Experts (MoE) paradigm, we use the features encoded by different expert $E_k$ to represent the query $x_q$ under various distinct modes.
\begin{equation}
q_k \;=\; E_k\,(u_q) \,, \qquad k=1,\ldots,K .
\end{equation}
\noindent where \(E_k:\mathbb{R}^{d}\!\to\!\mathbb{R}^{d^{'}}\) is the \(k\)-th expert projection, and \(q_k\in\mathbb{R}^{d^{'}}\) is the mode-specific representation by expert \(E_k\).

We employ a router \(R\) to produce a probability distribution over the \(K\) modes for the current query \(\pi_q = R(u_q)\).
\begin{equation}
\pi_q \in \Delta^{K}
\;\equiv\;
\big\{\pi \in \mathbb{R}^{K}_{\ge 0}\ \big|\ \textstyle\sum_{k=1}^{K}\pi_k = 1\big\}.
\end{equation}

Finally, we follow the mixture weights $\pi_q$ produced by the router $R$ to extract the information most relevant to the selected modes, yielding the query-side label-aware embedding $\tilde{u}_q \;=\; \sum_{k=1}^{K} \pi_{q,k}$.

\paragraph{Label-Aware Prompt Embeddings}

Then, we straightforward use the prompt label $L_i^p$ as an auxiliary cue and fuse it with the prompt image $I_i^p$ to obtain a joint image–label embedding $z_i = (z_i^I + z_i^L)$.

We instantiate a set of experts $\bar{E}$ on prompt side, symmetric to query side with independent parameters, to depict corresponding mode-specific embeddings for each prompt.
\begin{equation}
p_{i,k} \;=\; \bar{E}_k\,(z_i) \,, \qquad k=1,\ldots,K .
\end{equation}

\emph{Notably, our method adopts retrieval-based prompt selection. The mode-specific embeddings of database prompts $p_{i,k}$ can be preprocessed and cached.}

\paragraph{Extracting Query-Relevant Prompt Embeddings}

We use the query-adaptive router's output  \(\pi_q\) as target weights and combine the mode-wise prompt representations $p_{i,k}$ to extract the portion of query-relevant information  $\tilde{p}_{i\mid q}$.
\begin{equation}
\tilde{p}_{i\mid q} \;=\; \sum_{k=1}^{K} \pi_{q,k}\, p_{i,k} \,,
\end{equation}

\paragraph{Calculating Similarity and Select Best Prompt}

We finally match query-relevant prompt embeddings $\tilde{p}_{i\mid q}$ with label-aware query embedding $\tilde{u}_q$ using cosine similarity $\sigma(\tilde{u}_q,\tilde{p}_{i\mid q})=\langle \tilde{u}_q,\tilde{p}_{i\mid q}\rangle/(\|\tilde{u}_q\|_2\,\|\tilde{p}_{i\mid q}\|_2)$. And we select $i^{\star}=\arg\max_{i\in[N]} \sigma(\tilde{u}_q,\tilde{p}_{i\mid q})$ and then set $c_q^{\star}=(I_{i^{\star}}^{p},L_{i^{\star}}^{p})$ as the best prompt for VICL inference.

\subsection{Optimization Process of \modelname{}}
\label{subsec:training}

During optimization, we freeze the Vision Transformer feature extractor \(f\) throughout and train only the experts $E,\bar{E}$ and the router \(R\). 
Considering the two components play different roles, we adopt a decoupled two-step scheme in which each mini-batch undergoes two successive updates. 
In the first step, the experts $E,\bar{E}$ are optimized with a VICL performance objective while the mixture proportions \(\pi_q=R(u_q)\) are kept fixed, strengthening representations under different modes. In the second step, the experts are frozen and only the router \(R\) is updated using a prompt–query label-matching signal improving the alignment of the inferred mode mixture with the ground-truth query label. Additionally, we introduce an MoE-specific load-balancing loss encourages every expert to promotes even expert usage across queries.

\paragraph{Optimizing Mode-Specific Experts}

Retrieval-based prompt selection is typically optimized with a contrastive objective, and the query serves as the anchor. We first obtain label-agnostic features with the frozen encoder \(f\), namely \(u_q=f(x_q)\) and \(z_i^I=f(I_i^{p})\), then compute the similarity \(s_{\mathrm{ret}}(q,i)=\sigma(u_q,z_i^I)\) and keep the nearest neighbors to form a candidate pool:
\begin{equation}
\tilde{\mathcal{C}}_q \;=\; \operatorname{Top50}_{i\in[N]}\, s_{\mathrm{ret}}(q,i).
\end{equation}
We pair each candidate \((I_i^{p},L_i^{p})\in\tilde{\mathcal{C}}_q\) with the query \(x_q\) and run the MAE-VQGAN \cite{MAE-VQGAN} to obtain task scores \(\mathcal{H}_{\text{vp}}(q,i)=\mathcal{S}_{\text{task}}\!\big(\Psi(x_q;(I_i^{p},L_i^{p})),\,y_q\big)\). We select the best and worst \(5\) prompts as positives and negatives within \(\tilde{\mathcal{C}}_q\):
\begin{equation}
\mathcal{P}_q^v=\operatorname{Top}^{+}_{5}\{\,\mathcal{H}_{\text{vp}}(q,i)\,:\,i\in\tilde{\mathcal{C}}_q\},
\end{equation}
\begin{equation}
\mathcal{N}_q^v=\operatorname{Top}^{-}_{5}\{\,\mathcal{H}_{\text{vp}}(q,i)\,:\,i\in\tilde{\mathcal{C}}_q\}.
\end{equation}

We encode a mini-batch of queries together with their random selected positive and negative prompts $i_q^+ \in \mathcal{P}_q^v,i_q^- \in \mathcal{N}_q^v$ into the label-aware query embeddings \(\tilde{u}_q=\sum_{k=1}^{K}\pi_{q,k}q_k\) and the query-relevant prompt embeddings \(\tilde{p}_{i\mid q}=\sum_{k=1}^{K}\pi_{q,k}p_{i,k}\) as illustrated in \Cref{subsec:embeddings}. We use the cosine similarity \(s_{\mathrm{CL}}(q,i)=\sigma(\tilde{u}_q,\tilde{p}_{i\mid q})\) as similarity metric in contrastive objective.

Let the current mini-batch of queries be \(\mathcal{Q}_{\mathrm{mb}}\). For each \(q\in\mathcal{Q}_{\mathrm{mb}}\) define the denominator index set
\begin{equation}
\mathcal{D}_q^v \;=\; \\\bigcup_{q'\in\mathcal{Q}_{\mathrm{mb}}}\{i_{q'}^{+},i_{q'}^{-}\}.
\end{equation}

The contrastive loss is computed as
\begin{equation}
\mathcal{L}_{\mathrm{PG}}
\;=\;
-\, \frac{1}{|\mathcal{Q}_{\mathrm{mb}}|} \sum_{q\in\mathcal{Q}_{\mathrm{mb}}}
\log
\frac{ \exp\!\big(s_{\mathrm{CL}}(q,i_q^+)\big)}
     {\sum_{j\in\mathcal{D}_q^v} \exp\!\big(s_{\mathrm{CL}}(q,j)\big)}.
\label{eq:pg}
\end{equation}

We finally optimize the mode-specific experts $E,\bar{E}$ with the performance-guided contrastive loss $\mathcal{L}_\mathrm{PG}$.

\paragraph{Optimizing Expert Routing}

Given a query \(x_q\), the router $R$ outputs a probability vector \(\pi_q=R(u_q)\in\Delta^{K}\) that we regard as mixture weights over specific modes to estimate the unknown query label. 
To supervise this estimate, prompts with higher label compatibility are treated as positives. 
By pulling the query relevant embeddings of positives closer to the query, training calibrates \(\pi_q=R(u_q)\in\Delta^{K}\) toward label consistent modes.

We use the shortlist \(\tilde{\mathcal{C}}_q\) and the prompt–query label matching score \(\mathcal{S}_{\mathrm{task}}(L_i^{p},y_q)\) as the supervision signal. Positives and negatives are selected by \(\mathcal{S}_{\mathrm{task}}(L_i^{p},y_q)\)
\begin{equation}
\mathcal{P}_q^l=\operatorname{Top}^{+}_{5}\{\,\mathcal{S}_{\text{task}}(L_i^{p},y_q)\,:\,i\in\tilde{\mathcal{C}}_q\},
\end{equation}
\begin{equation}
\mathcal{N}_q^l=\operatorname{Top}^{-}_{5}\{\,\mathcal{S}_{\text{task}}(L_i^{p},y_q)\,:\,i\in\tilde{\mathcal{C}}_q\}.
\end{equation}

Subsequently we follow the same sampling protocol. For the current query \(x_q\), we sample one positive \(i_q^{+}\in\mathcal{P}_q^{l}\) and one negative \(i_q^{-}\in\mathcal{N}_q^{l}\). Then we transform them into their corresponding label-aware embeddings.
We define the batch-wise denominator index set
\begin{equation}
\mathcal{D}_q^l \;=\; \\\bigcup_{q'\in\mathcal{Q}_{\mathrm{mb}}}\{e_{q'}^{+},e_{q'}^{-}\}.
\end{equation}

The label-guided contrastive loss is computed as
\begin{equation}
\mathcal{L}_{\mathrm{LG}}
\;=\;
-\, \frac{1}{|\mathcal{Q}_{\mathrm{mb}}|} \sum_{q\in\mathcal{Q}_{\mathrm{mb}}}
\log
\frac{ \exp\!\big(s_{\mathrm{CL}}(q,e_q^+)\big)}
     {\sum_{j\in\mathcal{D}_q^l} \exp\!\big(s_{\mathrm{CL}}(q,j)\big)}.
\label{eq:lg}
\end{equation}

We also add a load-balancing objective for the router \(R\) to discourage expert under-utilization and encourage every mode to be selected. For a mini-batch \(\mathcal{Q}_{\mathrm{mb}}\) of size \(B\), we aggregate mixture weights into a batch distribution \(\bar{\pi}\) with components \(\bar{\pi}_k=\tfrac{1}{B}\sum_{q\in\mathcal{Q}_{\mathrm{mb}}}\pi_{q,k}\) and use the uniform target \(r_k=1/K\). The objective is expressed as 
\begin{equation}
\mathcal{L}_{\mathrm{LB}}
\;=\;
\mathrm{KL}\!\big(\bar{\pi}\,\|\,r\big)
\;=\;
\sum_{k=1}^{K}\bar{\pi}_k \log\!\Big(\frac{\bar{\pi}_k}{1/K}\Big).
\label{eq:lb}
\end{equation}

Finally, the router's learning objective can be shown as
\begin{equation}
    \mathcal{L}_R = \mathcal{L}_\mathrm{LG}+\mathcal{L}_\mathrm{LB}.
\label{eq:r}
\end{equation}

\section{Experiments} \label{sec:experiments}

\subsection{Downstream Tasks and Datasets}

To enable a fair and comprehensive comparison with current state-of-the-art methods~\cite{SupPR,PromptSelF,Partial2Global, RH-Partial2Global,aaai25}, we consider three widely adopted sub-tasks in visual in-context learning (VICL), each reflecting a fundamental visual ability: (i) \textbf{foreground segmentation}, which measures fine-grained image understanding; (ii) \textbf{single-object detection}, which evaluates spatial localization capability; and (iii) \textbf{image colorization}, which assesses generative reconstruction. Corresponding benchmark datasets are used for each task.

\textbf{Foreground Segmentation}. We adopt the $\text{Pascal-5}^{i}$~\cite{pascal-5i} dataset, which contains 20 categories in total. Each sample is a paired image and its corresponding segmentation mask. The dataset is partitioned into four folds, each consisting of five categories. The number of samples per fold ranges from 346 to 725. For training, we use 2286, 3425, 5583, and 2086 image–mask pairs for the four folds, respectively.
\textbf{Object Detection}. We adopt the Pascal VOC 2012 dataset~\cite{pascal-VOC}, which contains 20 object categories. Following standard practice, we use 612 image–annotation pairs for training. Each pair consists of an image and its corresponding bounding-box annotation for the target object.
\textbf{Colorization}. For the colorization task, we randomly sample 50k images from the 1.2M training set of ImageNet-1K ILSVRC2012~\cite{imagenet} to form the training split, and use the original validation set as the test split. Each pair consists of a grayscale input image and its corresponding colorized image.

\begin{table*}[t]
\centering
\caption{Comparison of \modelname{} with existing methods across three VICL tasks, including foreground segmentation, single-object detection, and colorization. The best performance values are \textbf{boldfaced}, and the second-best ones are \textit{italicized} for clarity.}

\resizebox{0.88\linewidth}{!}{
\begin{tabular}{l|c|ccccc|c|c}
\hline

\rowcolor{gray!8}
 &  & \multicolumn{5}{c}{\textbf{Seg. (mIoU) $\uparrow$} } &  &  \\
\rowcolor{gray!8}
\raisebox{0.5\normalbaselineskip}{\textbf{Prompt Selection Method}} & \raisebox{0.5\normalbaselineskip}{\textbf{Ref.}} & \textbf{Fold-0} & \textbf{Fold-1} & \textbf{Fold-2} & \textbf{Fold-3} & \textbf{AVG} & \raisebox{0.5\normalbaselineskip}{\textbf{Det. (mIoU) $\uparrow$}} & \raisebox{0.5\normalbaselineskip}{\textbf{Col. (MSE) $\downarrow$}}\\ \hline \hline
Random~\cite{MAE-VQGAN}                      & NIPS 2022  & 28.66 & 30.21 & 27.81 & 23.55 & 27.56 & 25.45 & 0.67 \\
UnsupPR~\cite{SupPR}                      & NIPS 2023  & 34.75 & 35.92 & 32.41 & 31.16 & 33.56 & 26.84 & 0.63 \\
SupPR~\cite{SupPR}                      & NIPS 2023  & 37.08 & 38.43 & 34.40 & 32.32 & 35.56 & 28.22 & 0.63 \\
\citet{aaai25} & AAAI 2025 & 36.86 & 42.22 & 37.11 & 30.84 & 36.76 & 28.25 & 0.62\\
Partial2Global~\cite{Partial2Global}              & NIPS 2024  & 38.81 & 41.54 & 37.25 & 36.01 & 38.40 & 30.66 & \textit{0.58} \\
RH-Partial2Global~\cite{RH-Partial2Global}           & NIPS 2025  & \textit{39.25} & \textit{42.15} & \textit{38.06} & \textit{36.60} & \textit{39.02} & 3\textit{0.94} & \textbf{0.56} \\
\rowcolor{yellow!12} \modelname{} \textbf{(Ours)}                 &    CVPR 2026 & \textbf{41.92} & \textbf{46.27} & \textbf{39.63} & \textbf{37.63} & \textbf{41.36} & \textbf{32.01} & \text{0.60} \\ \hline 
UnsupPR w/ voting~\cite{SupPR}    & NIPS 2023  & 41.07 & 41.32 & 38.14 & 36.44 & 39.24 & --- & --- \\
Prompt-SelF~\cite{PromptSelF}                 & TIP 2025  & 42.48 & 43.34 & 39.76 & 38.50 & 41.02 & 29.83 &  --- \\
Partial2Global w/ voting~\cite{Partial2Global}    & NIPS 2024  & \textit{43.23} & 45.50 & \textit{41.79} & \textit{40.22} & \textit{42.69} & 32.52 &    --- \\
RH-Partial2Global w/ voting~\cite{RH-Partial2Global} & NIPS 2025  & \textbf{43.53} & 45.88 & \textbf{41.99} & \textbf{40.90} & \textbf{43.08} & \textit{33.28} &    --- \\
\rowcolor{yellow!12} \modelname{} \textbf{ w/ voting (Ours)}       &    CVPR 2026 & \text{42.81} & \textbf{47.44} & \text{40.52} & 38.30 & \text{42.27} & \textbf{34.64} &   --- \\ \hline \hline
\end{tabular}
}

\label{tab:main_results}
\end{table*}

\subsection{Implementation Details}

\paragraph{Model Architecture}
The vision encoder from CLIP~\cite{CLIP} is employed to extract visual representations, and the parameters of the CLIP encoder are kept frozen throughout training. The mode-specific experts are implemented as lightweight MLP-based~\cite{mlp} networks. 
The MAE-VQGAN~\cite{MAE-VQGAN} model is adopted as the VICL backbone, and we use the checkpoint-3400 version in all experiments. Number of mode-specific experts $K$ is 10 in standard settings.

\paragraph{Training Details}
The trainable parameters are updated using the SGD optimizer with a learning rate of 0.005 and a batch size of 64. All the tasks are trained under settings for 200 epochs on a single NVIDIA A100 GPU (40GB). Optimization alternates between expert and router steps, and each mini-batch performs two successive updates.

\paragraph{Metrics}
For the segmentation and detection tasks, performance is evaluated using mean Intersection over Union (mIoU). For the image colorization task, Mean Squared Error (MSE) is adopted as the evaluation metric.

\subsection{Comparison with State-of-the-arts}  \label{subsec:sota_experiments}
\subsubsection{Baselines}

We conduct a comprehensive comparison of diverse and competitive representative methods for prompt selection in VICL, including Random~\cite{MAE-VQGAN}, UnsupPR~\cite{SupPR}, SupPR~\cite{SupPR}, Prompt\mbox{-}SelF~\cite{PromptSelF}, Partial2Global~\cite{Partial2Global}, \citet{aaai25} and RH\mbox{-}Partial2Global~\cite{RH-Partial2Global}, together with their corresponding voting variants described in Section~\ref{subsec:related_work_a}.

\subsubsection{Quantitative Results under Standard Protocols}

The quantitative results of the main experiments are presented in Table \ref{tab:main_results}. We design two experimental settings, depending on whether the voting strategy is applied. Without voting setting, our \modelname{} achieves the best performance across all three downstream tasks among retrieval-based methods.
Even against the strongest prompt selection baseline, the rerank-based RH-Partial2Global, retrieval-based \modelname{} delivers gains of $6.00\%$, $3.46\%$ on segmentation and detection respectively. 
Under the voting configuration, \modelname{} attains segmentation performance that is essentially on par with the rerank-based RH-Partial2Global, while still improving detection by $4.09\%$.
\emph{Experimental results demonstrate that label information is a crucial auxiliary signal for prompt selection, leading to marked improvements.}

\begin{table}[t]
\centering
\caption{Transferability across folds for \modelname{}, SupPR, and Partial2Global on the segmentation benchmark. Each method is trained on one fold and evaluated on the remaining folds.}
\label{tab:crossfold-vertical}
\resizebox{\linewidth}{!}{
\begin{tabular}{l|l|ccccc}
\toprule
\rowcolor{gray!8}
& & \multicolumn{5}{c}{\textbf{Target}} \\
\rowcolor{gray!8}
\raisebox{0.5\normalbaselineskip}{\textbf{Method}} & \raisebox{0.5\normalbaselineskip}{\textbf{Source}} & \textbf{Fold-0} & \textbf{Fold-1} & \textbf{Fold-2} & \textbf{Fold-3} & \textbf{AVG} \\
\midrule
\hline
\multirow{4}{*}{\shortstack{SupPR} }
  & Fold-0 & --- & 35.46 & 32.44 & 30.95 & 32.95 \\
  & Fold-1 & 34.92 & --- & 32.96 & 31.03 & 32.97 \\
  & Fold-2 & 34.71 & 36.48 & --- & 30.08 & 33.76 \\
  & Fold-3 & 34.01 & 35.83 & 32.15 & --- & 34.00 \\
\midrule
\multirow{4}{*}{\shortstack{Partial2Global}}
  & Fold-0 & --- & 36.38 & 32.63 & 30.90 & 33.30 \\
  & Fold-1 & 35.74 & --- & 32.94 & 31.32 & 33.33 \\
  & Fold-2 & 34.16 & 36.16 & --- & 30.44 & 33.59 \\
  & Fold-3 & 34.28 & 35.93 & 32.98 & --- & 34.40 \\
\midrule
\rowcolor{yellow!12}
  & Fold-0 & --- & \textbf{43.82} & \textbf{37.25} & \textbf{34.35} & \textbf{38.47} \\
\rowcolor{yellow!12}  & Fold-1 & \textbf{39.27} & --- & \textbf{37.06} & \textbf{33.33} & \textbf{36.55} \\
\rowcolor{yellow!12}  & Fold-2 & \textbf{39.78} & \textbf{43.09} & --- & \textbf{32.66} & \textbf{38.51} \\
\rowcolor{yellow!12} \multirow{-4}{*}{\modelname{} \textbf{(Ours)}}  & Fold-3 & \textbf{39.39} & \textbf{43.42} & \textbf{36.82} & --- & \textbf{39.87} \\
\hline
\bottomrule
\end{tabular}
}
\end{table}

\subsubsection{Cross-Fold Transferability}

Transferability across folds is a key criterion for prompt selection in VICL. We train \modelname{}, retrieval-based SupPR \cite{SupPR}, and rerank-based Partial2Global \cite{Partial2Global} on each fold of the segmentation benchmark. The trained retriever is then applied to the remaining folds to score candidates or to produce embeddings for nearest-neighbor selection. Results in \Cref{tab:crossfold-vertical} show that \modelname{} attains the strongest cross-fold transfer, surpassing SupPR by $14.8\%$ and Partial2Global by $13.9\%$. We attribute the substantial gains to \emph{distinct experts capturing modes that are finer grained than category information}, and the router adaptively selects, for each query, the most critical components to extract.

\begin{table*}[t]
\centering
\caption{Ablation study on segmentation and detection across different configurations.}
\label{tab:ablation_cvpr_style}
\resizebox{0.86\linewidth}{!}{
\begin{tabular}{lllcccccc}
\hline
\rowcolor{gray!8}
& &  &
\multicolumn{5}{c}{\textbf{Seg. (mIoU $\uparrow$)}} & \\
\rowcolor{gray!8}
\multirow{-2}{*}{\textbf{Type}} & \multirow{-2}{*}{\textbf{ID}} & \multirow{-2}{*}{\textbf{Method}} & \textbf{Fold-0} & \textbf{Fold-1} & \textbf{Fold-2} & \textbf{Fold-3} & \textbf{AVG} & \multirow{-2}{*}{\textbf{Det. (mIoU $\uparrow$)}}  \\
\hline \midrule
\rowcolor{yellow!12}
\textbf{Full Model } & (0) & \modelname{} \textbf{(Ours)}            & 41.92 & 46.27 & 39.63 & 37.63 & 41.36   & 32.01 \\
\midrule 
& (1) & w/o Router             & 38.27 & 43.31 & 36.23 & 34.99 & 38.20   & 29.69 \\
 
\multirow{-2}{*}{Architectural Components} & (2) & w/o Prompt Label       & 39.14 & 44.24 & 37.82 & 35.56 & 39.19   & 30.94 \\
\midrule 
Feature Extractor Substitution & (3) & CLIP $\rightarrow$ DINOv2 & 41.51 & 46.08 & 39.93 & 38.05 & 41.39 & 32.06 \\
\midrule 
Optimization Strategy & (4) & Single-Stage Training  & 40.16 & 44.70 & 38.34 & 36.22 & 39.86  & 31.21 \\
\midrule
& (5) & w/o $\mathcal{L}_\mathrm{PG}$           & 36.15 & 37.47 & 33.91 & 32.68 & 35.05 & 27.30 \\
& (6) & w/o $\mathcal{L}_\mathrm{LG}$           & 40.26 & 44.53 & 38.10 & 35.77 & 39.67  & 30.14 \\
\multirow{-3}{*}{Loss Function Ablation } & (7) & w/o $\mathcal{L}_\mathrm{LB}$           & 40.49 & 44.95 & 38.74 & 37.25 & 40.36 & 31.43 \\
\hline \hline
\end{tabular}
}
\end{table*}

\begin{figure*}[t]
    \centering
    \includegraphics[width=0.95\linewidth]{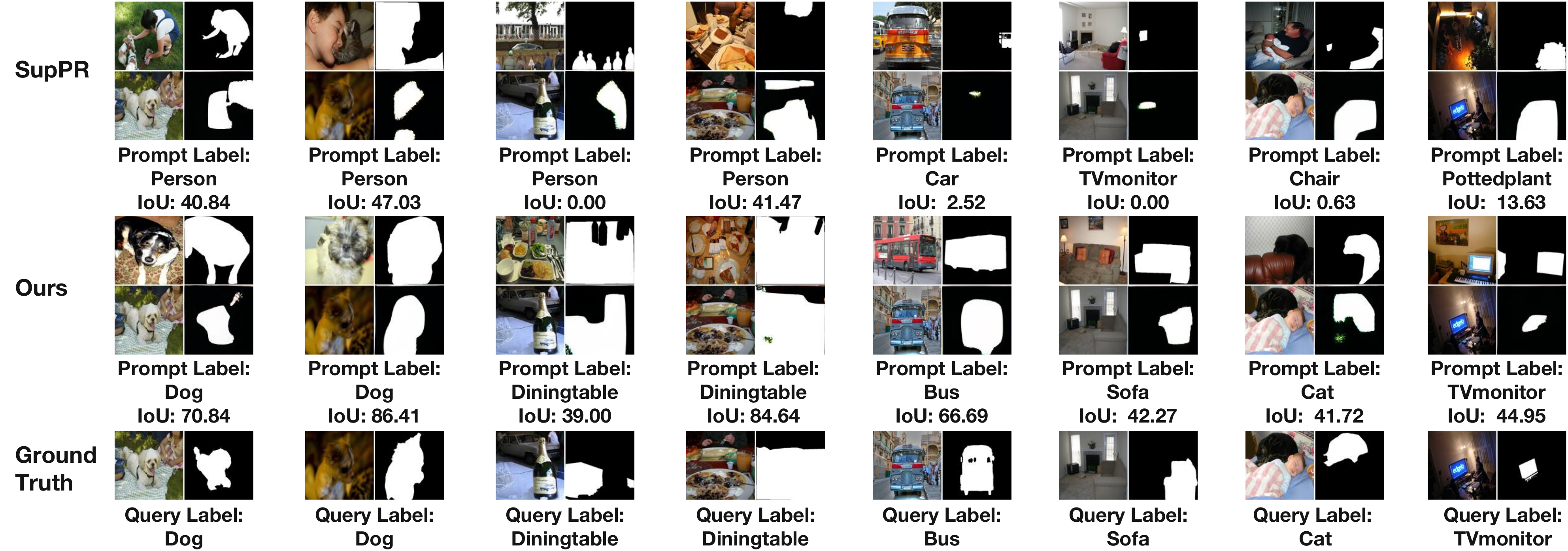}
    \caption{Qualitative visualization comparing \modelname{} (label-aware prompt retrieval) with SupPR (label-agnostic prompt retrieval). \modelname{} consistently retrieves label compatible prompts and yields visibly more accurate VICL predictions.}
    \label{fig:case}
\end{figure*}

\subsection{Ablation Analyses}  \label{subsec:ablation_study}
To comprehensively assess the contribution of each component, we construct a suite of ablated variants and assign a unique ID for clear reference. The results are summarized in \Cref{tab:ablation_cvpr_style}, where \textbf{Variant (0)} denotes our default configuration.

\subsubsection{Effectiveness of Label-Aware Embeddings}

We separately assess the query side and the prompt side. \textbf{Variant (1)} disables the router by replacing the mixture with a uniform distribution, which averages expert outputs irrespective of the query. \textbf{Variant (2)} removes explicit prompt labels when forming joint encodings by replacing with the image-only representation. Results in \Cref{tab:ablation_cvpr_style} show clear drops on both segmentation and detection. These findings indicate that explicit label injection on the prompt side and query conditioning via an estimated implicit label are both necessary, \emph{working together to align label-aware prompt embeddings with the query and achieve label consistency.} 

\subsubsection{Cross Different Feature Extractor}

To assess generality across feature extractors, we introduce \textbf{Variant (3)}, which replaces the encoder with DINOv2 \cite{oquab2023dinov2} and reevaluates \modelname{} under the same protocol. Comparable experiments have been reported for SupPR \cite{SupPR} and Partial2Global \cite{Partial2Global}.
\modelname{} remains the top performer across tasks and feature extractors, which confirms its effectiveness independent of the encoder. The results further show that DINOv2 features do not necessarily yield stronger retrieval than CLIP features. \emph{We attribute this to a gap between generic visual representations used for retrieval and the signals that best helps VICL prediction.} Bridging this gap is a promising direction for VICL retrieval.

\subsubsection{Effectiveness of Alternating Optimization}

Instead of the decoupled two step scheme, we design \textbf{Variant (4)} that adopts joint training and updates the experts and the router simultaneously under a unified objective \(\mathcal{L}_{\mathrm{J}}=\mathcal{L}_{\mathrm{PG}}+\mathcal{L}_{\mathrm{LG}}+\mathcal{L}_{\mathrm{LB}}\). Under identical settings, joint training yields lower accuracy on both segmentation and detection, and the loss shows larger fluctuations with slower convergence. These observations indicate that \emph{alternating optimization is more stable and effective}, with experts updated by the performance-guided objective \(\mathcal{L}_{\mathrm{PG}}\) and the router updated by the label-guided objective \(\mathcal{L}_{\mathrm{LG}}+\mathcal{L}_{\mathrm{LB}}\).

\subsubsection{Effectiveness of Learning Objectives}

We ablate each learning objective in isolation. \textbf{Variant (5)} replaces the expert step’s performance-guided objective \(\mathcal{L}_{\mathrm{PG}}\) with the label-guided objective \(\mathcal{L}_{\mathrm{LG}}\). \textbf{Variant (6)} replaces the router step’s label-guided objective \(\mathcal{L}_{\mathrm{LG}}\) with the performance guided objective \(\mathcal{L}_{\mathrm{PG}}\). \textbf{Variant (7)} removes the load balancing objective \(\mathcal{L}_{\mathrm{LB}}\).
Results show a large degradation when \(\mathcal{L}_{\mathrm{PG}}\) is removed. The accuracy falls to just above the unsupervised UnsupPR \cite{SupPR} baseline and remains below SupPR \cite{SupPR}. \emph{This indicates \(\mathcal{L}_{\mathrm{PG}}\) provides the primary supervision for learning an effective retriever.} Removing \(\mathcal{L}_{\mathrm{LG}}\) also yields a consistent drop, which confirms \emph{label guidance is an important auxiliary signal for selecting soft implicit query label}. Eliminating \(\mathcal{L}_{\mathrm{LB}}\) leads to a lesser reduction, which \emph{we attribute to expert insufficient utilization among modes.} These objectives are complementary and together deliver best performance of \modelname{}.

\subsection{More Analyses}

\subsubsection{Exploring the Role of Labels in Prompt Retrieval}

As illustrated in \Cref{fig:case}, we present several toy cases that expose a typical failure of label-agnostic retrieval.
\emph{SupPR often retrieves prompts whose images resemble the query while the labels disagree, sometimes even tagging categories absent from the query}, which degrades VICL inference. \emph{\modelname{} incorporates labels during retrieval and enforces image–label consistency}, yielding prompts that align with the query and produce stronger predictions. This underscores the importance of integrating labels into the retrieval pipeline.

\begin{figure}[t]
    \centering
    \includegraphics[width=0.92\linewidth]{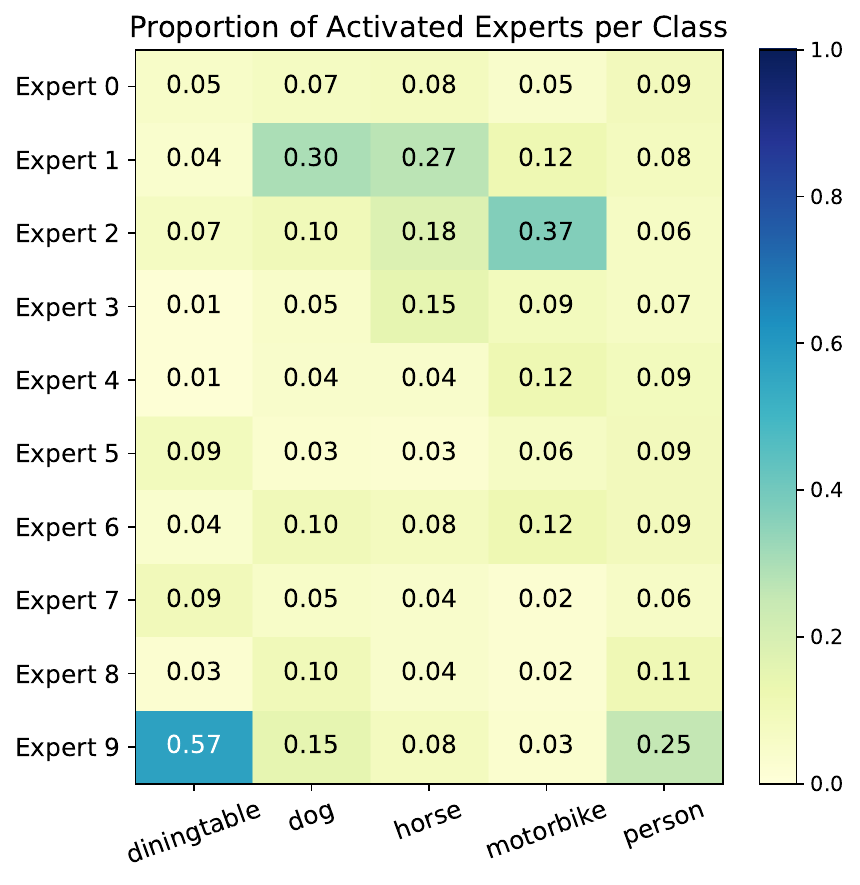}
    \caption{Proportion of each mode-specific expert selected under each category, visualized as a heatmap.}
    \label{fig:expert_class}
    \vspace{-1.5em}
\end{figure}

\subsubsection{Mode-Specific Experts Activation Proportions} \label{subsec:mode-activation}

To examine whether different experts have learned distinct modes and to verify that the router effectively selects label-aware information, we use the Pascal-$5^\mathrm{i}$~\cite{pascal-5i} dataset with the ``fold 2" split. On its validation set we measure, for five categories, the proportion with which each mode is activated, as shown in \Cref{fig:expert_class}. The activation ratios vary substantially across categories, and the modes emphasized by different categories are not the same, which aligns with our design goal. In addition, the classes ``dog" and ``horse" exhibit similar activation modes, indicating semantic affinity and suggesting that \emph{the learned modes capture structure that is finer grained than the category information.}

\section{Conclusions}
\label{sec:conclusion}

In this paper, we presented \modelname{}, a \textbf{l}abel-\textbf{a}ware \textbf{p}rompt \textbf{r}etrieval framework for VICL that treats labels as an important auxiliary signals. 
\modelname{} injects prompt labels to form joint representations.
It employs a mixture of experts to model mode specific embeddings on both the query and prompt sides, and a query conditioned router sets the mixture weights to obtain label-aware embeddings that promote label consistency.
Since the experts and the router play different roles, we optimize them with alternating training.
Extensive experiments on foreground segmentation, single object detection, and colorization show consistent gains. \modelname{} achieves generalizes across feature extractors, and transfers reliably under cross-fold. 
These results indicate that bringing labels into the prompt selection is an effective principle for VICL, which we hope can inspire further research.

\paragraph{Acknowledgments}
We sincerely thank the anonymous reviewers and chairs for their efforts and constructive suggestions, which have greatly helped us improve the manuscript. 
This work is supported in part by the National Natural Science Foundation of China under grants 624B2088, 62576122, 62571298, 62301189.
{
    \small
    \bibliographystyle{ieeenat_fullname}
    \bibliography{main}
}

\end{document}